\pdfoutput=1
\documentclass[letterpaper, 10 pt, conference]{ieeeconf}  % Comment this line out if you need a4paper

\IEEEoverridecommandlockouts                              % This command is only needed if 
\usepackage{graphics} % for pdf, bitmapped graphics files
\usepackage{epsfig} % for postscript graphics files
\usepackage{float}
\usepackage{times} % assumes new font selection scheme installed
\usepackage{amsmath} % assumes amsmath package installed
\usepackage{amssymb}  % assumes amsmath package installed
\usepackage{subcaption}
\usepackage[hidelinks]{hyperref}
\usepackage[table]{xcolor}
\usepackage{todonotes}
\usepackage{siunitx}
\usepackage{multirow} 
\usepackage{cite}
\usepackage{siunitx} % Add this to your LaTeX preamble for number alignment
\usepackage{pgfplots}

\usepackage[utf8]{inputenc} % allow utf-8 input
\usepackage[T1]{fontenc}    % use 8-bit T1 fonts
\usepackage{hyperref}       % hyperlinks
\usepackage{url}            % simple URL typesetting
\usepackage{booktabs}       % professional-quality tables
\usepackage{amsfonts}       % blackboard math symbols
\usepackage{nicefrac}       % compact symbols for 1/2, etc.
\usepackage{microtype}      % microtypography
\usepackage{xcolor}         % colors

\usepackage{amsmath,amsfonts,bm}

\def\eqref#1{equation~\ref{#1}}
\def\1{\bm{1}}

\DeclareMathAlphabet{\mathsfit}{\encodingdefault}{\sfdefault}{m}{sl}
\SetMathAlphabet{\mathsfit}{bold}{\encodingdefault}{\sfdefault}{bx}{n}

\usepackage{graphicx} % For \resizebox
\usepackage{tabularx} % For tabularx environment
\usepackage[acronym]{glossaries}
\usepackage{subcaption}
\usepackage{algorithm}
\usepackage{algorithmic}
\pgfplotsset{compat=1.18}
\usepackage{colortbl} % Required for coloring table cells
\usepackage{wrapfig}
\usepackage{lipsum} % For dummy text
\usepackage{amssymb}

\usepackage{tikz}
\usetikzlibrary{positioning, arrows, automata, matrix}
\usetikzlibrary{external}
\usetikzlibrary{backgrounds, calc}
\usetikzlibrary{spy}
\usetikzlibrary{fit}
\usetikzlibrary{arrows.meta}
\usetikzlibrary{automata,arrows,shapes.geometric,shapes.misc}

\usepackage{graphicx}
\usepackage{amsfonts}

\usepackage[automake=immediate,acronym]{glossaries-extra}
\makeglossaries
\usepackage{xspace}

\pgfplotsset{compat=1.17}
\usetikzlibrary{pgfplots.groupplots}
\title{\Large\bfseries%
Beyond Visuals: Investigating Force Feedback in Extended Reality for Robot Data Collection
}

\author{Xueyin Li$^{1*}$, Xinkai Jiang$^{1*}$, Philipp Dahlinger$^1$, Gerhard Neumann$^1$, Rudolf Lioutikov$^1$
\thanks{$^1$Intuitive Robots Lab, Karlsruhe Institute of Technology, Germany.}
\thanks{*Xueyin Li and Xinkai Jiang contributed equally to this work.}
}

\setabbreviationstyle[acronym]{long-short}

\glsdisablehyper
\newacronym{method}{VDD}{Variational Diffusion Distillation}
\newacronym{moe}{MoE}{Mixture of Experts}
\newacronym{lfd}{LfD}{Learning from Human Demonstration}
\newacronym{sota}{SoTA}{state-of-the-art}
\newacronym{beso1holder}{beso-1}{beso-1step}
\newacronym{ddpm1holder}{ddpm-1}{ddpm-1step}
\newacronym{besoholder}{beso}{beso-16steps}
\newacronym{ddpmholder}{ddpm}{ddpm-16steps}
\newacronym{consistency}{CD}{Consistency Distillation}

\newacronym{RA-L}{RA-L}{Robotics and Automation Letters}

\newacronym[longplural=degrees of freedom]{DoF}{DoF}{degree of freedom}
\newacronym{POEMPEL}{POEMPEL}{Plastic Over-Engineered Movement Primitive End-effector Link}
\newacronym{FAMP}{FA-ProDMP}{Force-Aware \Glsxtrshort{ProDMP}}
\newacronym{MP}{MP}{Movement Primitive}
\newacronym{DMP}{DMP}{Dynamic \Glsfmtlong{MP}}
\newacronym{ProMP}{ProMP}{Probabilistic \Glsfmtlong{MP}}
\newacronym{ProDMP}{ProDMP}{Probabilistic \Glsfmtlong{DMP}}
\newacronym{PiH}{PiH}{peg-in-hole}
\begin{document}

\maketitle
\thispagestyle{empty}
\pagestyle{empty}

%%%%%%%%%%%%%%%%%%%%%%%%%%%%%%%%%%%%%%%%%%%%%%%%%%%%%%%%%%%%%%%%%%%%%%%%%%%%%%%%
\glsunsetall % manual acronyms
\begin{abstract}
% With providing additional sensory information, force feedback has been proposed to enhance Human-Robot Interaction, its influence on Learning from Demonstration remains unclear.
% In this study, we investigate the impact of introducing force feedback into two fundamental interaction interfaces for Learning from Demonstration. We conduct a user study spanning simple to complex precise operation tasks, evaluating task performance, user experience, and the impact of force feedback on data quality for Imitation Learning. 
% Our results show that force feedback can provide significant benefits in challenging scenarios, improve user experience, and potentially improve the quality of demonstration data. However, the extent of these improvements depends on the specific interaction method and task complexity. This finding provides a new sight of how tasks influence the impact of force feedback and presents a more thorough examination of how force feedback affects data quality. 

% By providing contact-rich information, force feedback has been proved to enhance the user experience in Extended Reality (XR).
This work explores how force feedback affects various aspects of robot data collection within the Extended Reality (XR) setting.
Force feedback has been proved to enhance the user experience in Extended Reality (XR) by providing contact-rich information.
However, its impact on robot data collection has not received much attention in the robotics community.
This paper addresses this shortcoming by conducting an extensive user study on the effects of force feedback during data collection in XR.
We extended two XR-based robot control interfaces, Kinesthetic Teaching and Motion Controllers, with haptic feedback features.
% To address this shortcoming, 
% this paper haptic feedback features are implemented for two XR-based robot control interfaces: Kinesthetic Teaching and Motion Controllers.
The user study is conducted using manipulation tasks ranging from simple pick-place to complex peg assemble, requiring precise operations.
% To examine how force feedback affects robot data collection,
The evaluations show that force feedback enhances task performance and user experience, particularly in tasks requiring high-precision manipulation.
These improvements vary depending on the robot control interface and task complexity.
This paper provides new insights into how different factors influence the impact of force feedback.
% A comprehensive examination further shows how force feedback affects the data quality.

\end{abstract}

\glsresetall % reset acronym usage
\section{Introduction}
\label{sec:introduction}

Robot data collection
plays a crucial role in advancing robotics by enabling robots to acquire complex skills through human guidance \cite{pizero2024}. For effective learning, robots require large volumes of high-quality data, as the success of data-driven methods heavily depends on the quality and diversity of the training demonstrations \cite{argall2009survey}.

Extended Reality (XR) has emerged as a promising tool to facilitate the collection of such demonstrations. XR provides an immersive interface, allowing users to control robots in an intuitive manner. This technology further reduces setup time, making it an effective means for generating demonstrations. Despite its benefits, conventional XR-based robot data collection primarily relies on visual feedback, which may be insufficient for tasks demanding high precision.
%for task with visual occlusion

\begin{figure}[h]
    \centering
    \includegraphics[width=0.96\linewidth]{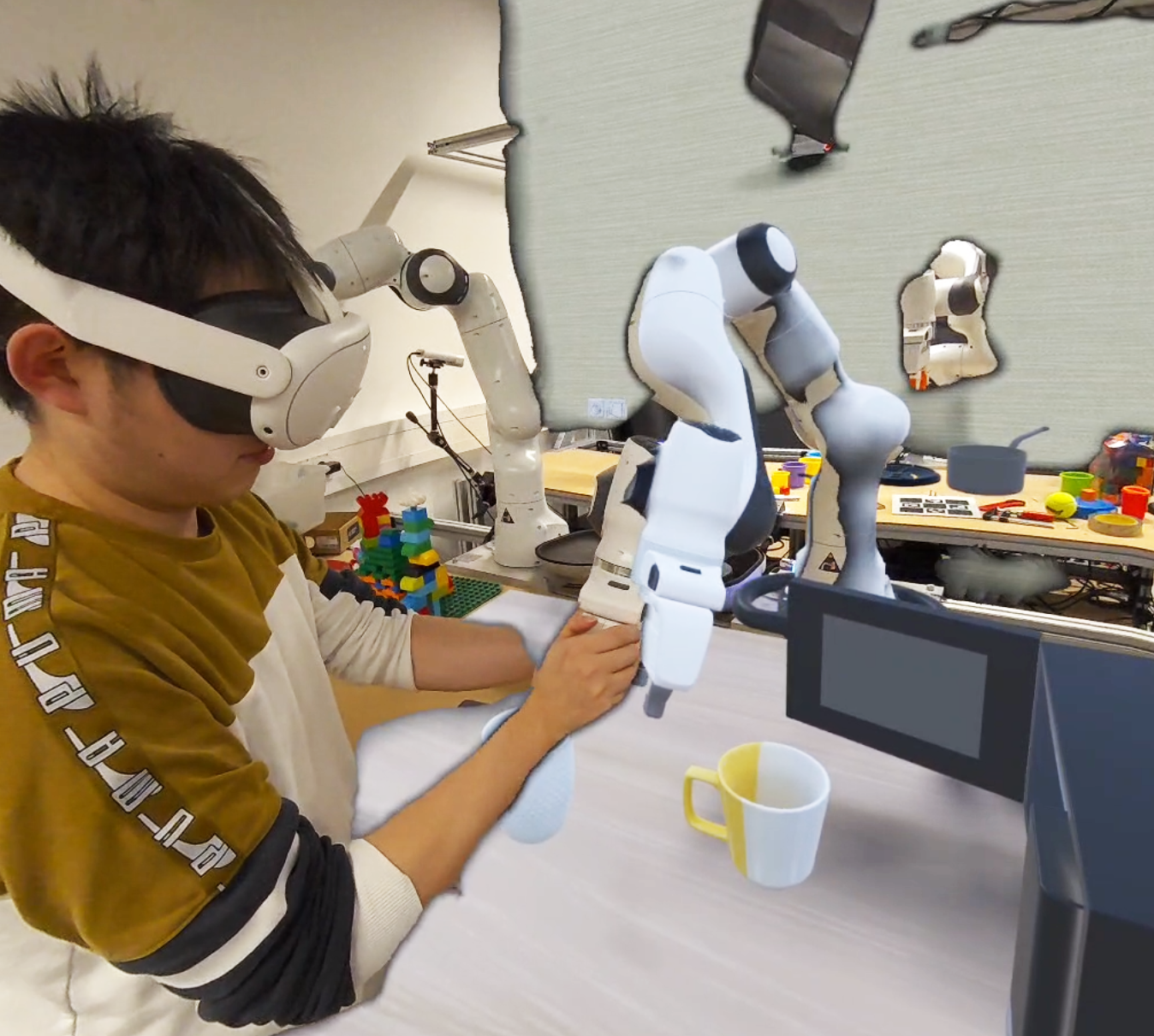} % Change to your actual image filename
    \caption{
Kinesthetic Teaching (KT) for data collection with force feedback. The virtual and real robots are aligned, ensuring that all movements of the real robot are accurately projected onto the virtual counterpart.
    }
    \label{fig:ueq_s_plot}
\end{figure}

Beyond visual information, haptic feedback is a promising complement to robot data collection. To facilitate tasks that demand high precision, force feedback has been introduced as an additional sensory channel, enriching teleoperation by providing tactile information. Force feedback refers to the use of actuators to generate resistive or assistive forces, allowing users to feel physical interactions with the virtual or remote environment \cite{overtoom2019haptic}.

However, the impact of force feedback in robot teleoperation and data collection remains underexplored.
The effectiveness of force feedback across different control paradigms is not yet fully understood, and its influence on task performance has not been systematically evaluated.
To address this shortcoming, this study investigates the role of force feedback in an XR-based teleoperation setup through the following hypotheses:

\textbf{Hypothesis 1 (H1):} Force feedback enhances the efficiency, effectiveness, and user experience of data collection.

\textbf{Hypothesis 2 (H2):} The influence of force feedback varies across different interfaces.

\textbf{Hypothesis 3 (H3):} The impact of force feedback on efficiency, effectiveness, and user experience varies across different task types.

To test these hypotheses, we conducted a systematic user study where participants controlled a virtual robot in simulation using different XR-based interfaces. The study focused on four interfaces, including two primary ones: Kinesthetic Teaching and Motion Controller, both with and without force feedback.
The combination of an XR environment with force feedback enabled Kinesthetic Teaching is novel to the best of our knowledge.
A total of 31 participants completed a series of tasks, providing demonstration data while interacting with each interface. Performance was evaluated using both objective metrics (task success rate and completion time) and subjective metrics (questionnaire assessing user experience).

The results of the study indicate that force feedback with Kinesthetic Teaching in XR is a powerful and intuitive approach for collecting demonstrations with high performance and good user experience. Interfaces incorporating force feedback demonstrated improved precision and stability, especially in tasks requiring fine motor control. Furthermore, the impact of force feedback varied across different control modalities and tasks, suggesting that its benefits are context-dependent.

In summary, this work makes two key contributions. First, it presents a comprehensive user study that systematically evaluates the impact of force feedback on robot teleoperation and data collection. Second, it introduces a force feedback-based Kinesthetic Teaching interface, which outperforms its non-feedback counterpart in terms of performance and user experience. These insights contribute to the development of more effective robot teleoperation systems and may inform future applications in real-world robotic control and teleoperation.

\section{Related Work}

\subsection{Force Feedback in Human-Robot Interaction}

Force perception is fundamental to how humans interact with the world, enabling us not only to engage with our environment but also to perceive these interactions simultaneously \cite{minogue2006haptics}. Force feedback has been shown to enhance position control, improve navigation and precision tasks, and reduce reliance on visual and auditory modalities \cite{sigrist2013augmented}. It also improves temporal accuracy \cite{boessenkool2012task} and enhances overall user comfort \cite{rahal2020caring}.
Recently, integrating force feedback into robot teleoperation interfaces has gained increasing attention\cite{bong2022force,patel2022haptic, lodige2024use}.
% Researchers have explored various methods to implement force feedback in teleoperation. 
While physical interactions with real robots allow operators to experience reaction forces, this feedback is absent in virtual environments and teleoperation settings. To address this limitation, different force feedback modalities have been introduced to improve interaction fidelity.
Some approaches use twin controllers with the same parameters and sensors as the robot to provide direct force feedback to the operator’s hand and wrist \cite{kim2023training} or industrial force feedback devices designed for robotics applications\cite{dombrowski2017interactive}.
Others employ force-feedback gloves with motor vibrations \cite{becker2024integrating} or integrate bimanual robot avatars with upper-body controllers to enhance immersion through haptic feedback \cite{lenz2023bimanual}.

In XR environments, vibration-based force feedback from controllers has been shown to enhance immersion, increase perceived performance, and reduce task difficulty \cite{brasen2019effects}. This approach is particularly beneficial for tasks requiring delicate manipulation, such as handling fragile objects \cite{cuan2024leveraging} or deformable materials \cite{becker2024integrating}.

Despite these advances, the impact of different force feedback modalities in human-robot interaction remains underexplored. By incorporating force feedback, interaction systems can become more intuitive and efficient, improving performance and user satisfaction across various applications. However, few studies have explored the influence of different force feedback modalities on tasks performance and user experience in HRI systems.

\subsection{XR in Robot Data Collection}

Extended Reality (XR), including Virtual Reality (VR), Augmented Reality (AR), and Mixed Reality (MR), combines real objects with virtual environments to enhance Human-Robot Interaction\cite{chuah2018and}. In recent years, XR has been widely applied in robotics, promoting the development of more intuitive and efficient control methods and interfaces\cite{wang2024towards}.
In robot teleoperation, XR enables operators to manipulate robots within virtual environments, thereby improving operational efficiency and reducing workload\cite{whitney2019comparing}. XR can intuitively convey visual information to operators and even provide multiple views, offering users an immersive experience\cite{wei2021multi}.

Numerous studies have demonstrated that XR devices can be used to control or interact with real robots\cite{suzuki2022augmented, hirschmanner2019virtual}. XR provides depth perception and intuitive visual feedback, allowing users to perceive spatial relationships more accurately, thereby significantly improving operational efficiency\cite{macias2020measuring}. Many studies also show that AR interfaces reduce task completion time, enhance operator performance, and are generally preferred over traditional interfaces\cite{wonsick2020systematic}. Compared to other display devices, VR headsets enable users to perform manipulation tasks faster, with lower perceived workload and higher usability\cite{whitney2019comparing}. Furthermore, research indicates that in XR environments, virtual robots can be as accurate and efficient as physical robots—or even more so\cite{han2023crossing}. This opens possibilities for teleoperation and simulation without the need for physical robot hardware, reducing research costs and increasing accessibility.

Previous studies have shown that using kinesthetic teaching (KT) interfaces and VR motion controller (MC) interfaces to control virtual robots in XR devices results in higher efficiency and effectiveness compared to other control interfaces\cite{jiang2024comprehensive}. The kinesthetic teaching interface allows operators to interact directly with the physical entity of the virtual robot, while the VR motion controller guides the virtual robot's end-effector through manual control.

To collect demonstration data in virtual environments using XR interfaces, these two interfaces (KT and MC) are adopted as the primary methods for controlling virtual robots. Leveraging the IRIS framework \cite{jiang2025irisimmersiverobotinteraction}, XR devices are integrated to provide intuitive visual feedback and incorporate different forms of force feedback.

% \section{Hypotheses}

% \textbf{
% Hypothesis 1 (H1): 
% Force feedback enhances the efficiency, effectiveness, and user experience of data collection.
% }

% \textbf{
% Hypothesis 2 (H2): 
% Different types of force feedback vary the impacts of the performance.
% }

% \textbf{
% Hypothesis 3 (H3): 
% The efficiency, effectiveness, and user experience impacts of introducing force feedback vary across different tasks.
% }

% \textcolor{red}{not sure if we need h4 for data quality evaluation}

% \textbf{
% Hypothesis 4 (H4): 
% Force feedback enhances the efficiency, effectiveness, and user experience of data collection.
% }

\section{Experiment}

\subsection{System Design}
% \subsubsection{XR Platform and Physics Simulator}
% Our XR framework, as outlined in \cite{jiang2025irisimmersiverobotinteraction}, uses the Meta Quest 3 to immerse users in a virtual environment, with the simulation continuously sending updates to the headset, including the position and rotation of each object. The simulation scene is highly adjustable, offering greater flexibility than real-world environments. For physics simulation, we use Mujoco \cite{mujoco}, and data loggers are implemented to record the state information of the virtual robot and objects, such as position, velocity, acceleration, and orientation.
\subsubsection{XR Platform and Physics Simulator}

Our XR framework is based on \cite{jiang2025irisimmersiverobotinteraction} and utilizes the Meta Quest 3 as the XR headset.
This framework projects scenes from the simulation into the XR headset.
The simulation runs on a PC, with the headset connected via a Wi-Fi router.
The XR framework transmits the entire scene, including all objects, meshes, materials, and textures, to the headset, generating an identical Unity scene. 
To ensure synchronization, the simulation continuously updates the headset with scene modifications, including the position and rotation of each object.

The simulation scene is highly adjustable, offering greater flexibility than real-world environments. This work uses Mujoco \cite{mujoco} as simulatior,
and data loggers are implemented to record the state information of the virtual robot and objects,
such as position, velocity, acceleration, and orientation.

% \subsubsection{Physics Simulator}
% Simulation scene can be easily adjusted, offering greater flexibility compared to real-world environments.
% This work uses Mujoco as physical simulator,
% and robot tasks and controllers was built by SimulationFramework.
% Additional data loggers were implemented to record the state information of the virtual robot and objects, including their position, velocity, acceleration, and orientation.

% real robot?Franka Emik Panda 

% \subsubsection{Initial Interfaces}
% kt
% mc

\begin{figure}[h]
    \centering
    \begin{minipage}{0.48\textwidth}
        \centering
        \subfloat[MCV]{\includegraphics[width=0.45\textwidth]{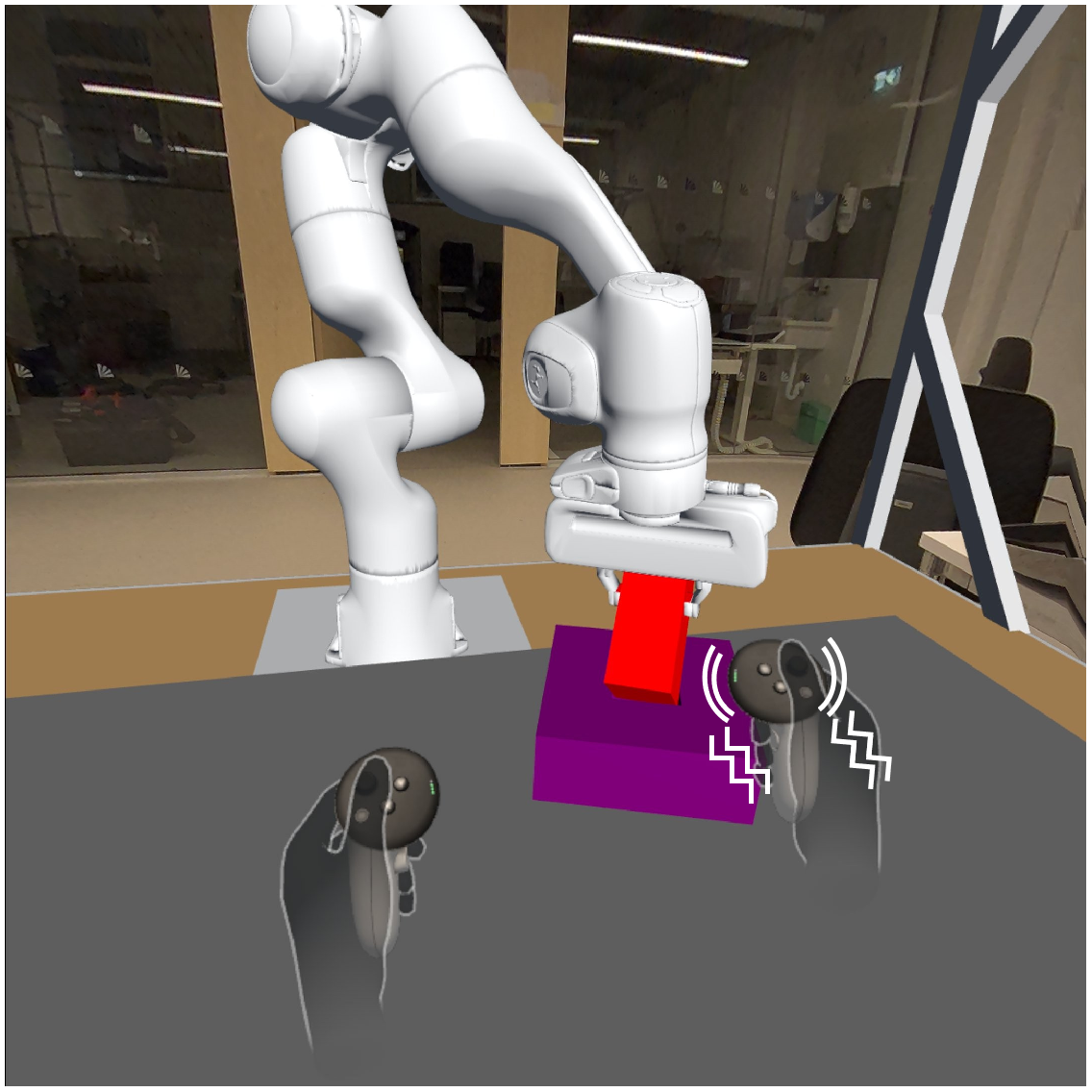}}
        \subfloat[KTF]{\includegraphics[width=0.45\textwidth]{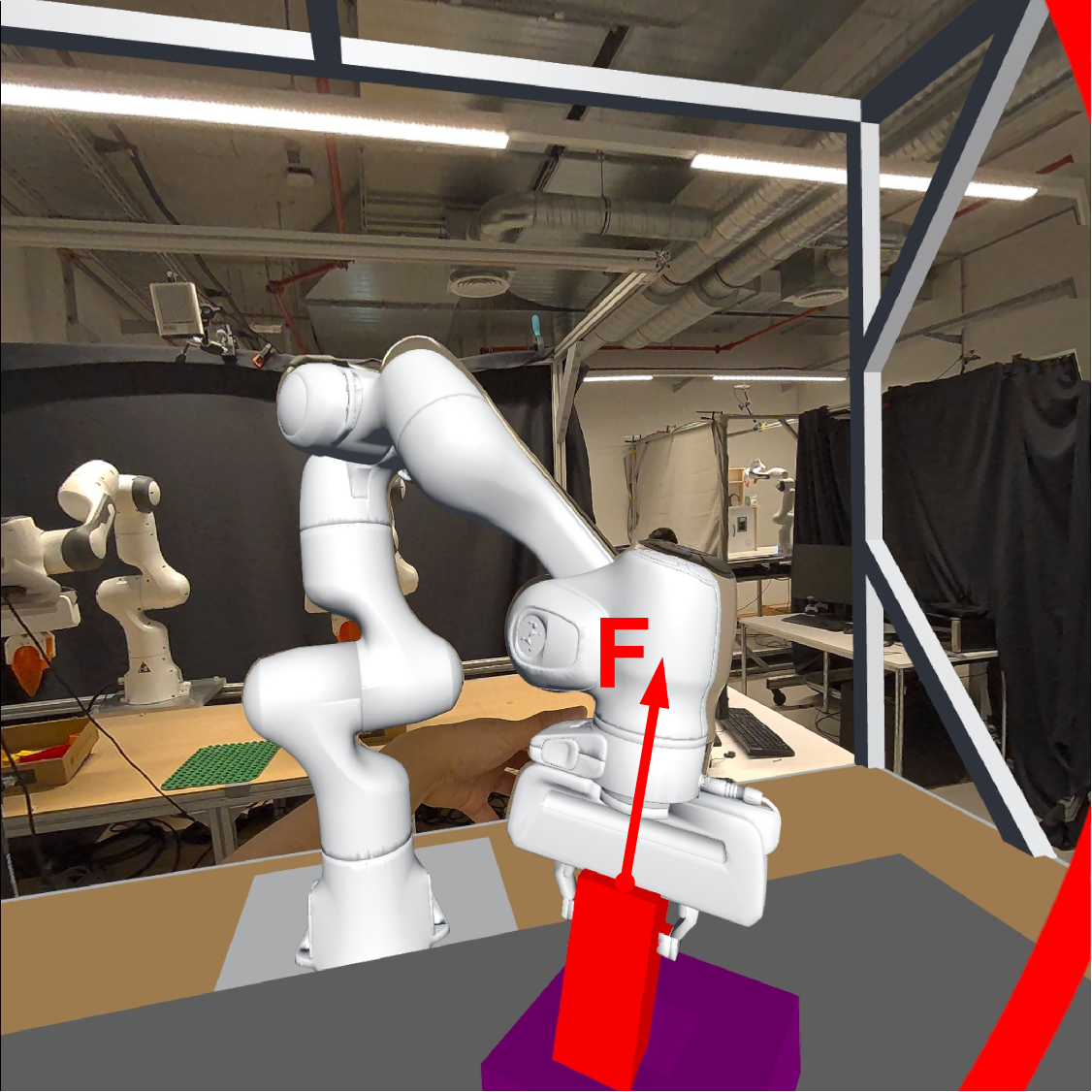}}
        \caption{Two types of robot control interfaces with force feedback. The force can be perceived either through the vibration of the motion controller or the applied force from the real robot. These images are captured from the perspective of XR headsets.}
        \label{fig:interface_comparison}
    \end{minipage}
\end{figure}

\subsubsection{Robot Control Interface}
In data collection tasks, robot control interfaces are used to operate the robot in both simulated and real-world environments.
Kinesthetic Teaching (KT) and Motion Controllers (MC) are among the most commonly used interfaces, as they provide intuitive and effective control.
KT is an intuitive interface that enables users to control robots by physically guiding their movements. In contrast, MC employ inverse kinematics to manipulate robots in Cartesian space, where the robot's end effector is controlled via the controller's trigger.
This study focuses on these two interfaces to investigate the impact of force feedback on their performance.

\subsubsection{Motion Controller with Vibration Feedback (MCV)}
% mcv
% Based on the initial MC interface, implicit force feedback is introduced through vibration generated by the Touch Plus controllers under specific conditions. In the simulation, contact forces between the gripper’s fingertips and the object, as well as collisions between manipulated objects and the environment, are detected. If the gripper touches an object without closing,  the Touch Plus controller provides continuous vibration. When the gripper attempts to grasp and maintain a hold, a short vibration is provided as a cue, after which no further feedback occurs while the object remains held. Upon releasing the object, new vibration feedback is reactivated upon contact. Additionally, if the gripped object collides with another object in the environment, vibration feedback persists until the contact ends.

Building on the initial MC interface, force feedback is provided through vibration from the Touch Plus controllers under specific conditions.
In the simulation, contact forces between the gripper’s fingertips and objects, as well as collisions with the environment, are detected. If the gripper touches an object without closing, continuous vibration is applied. A short vibration cue is given when the gripper grasps an object, with no further feedback while holding it. %Vibration reactivates upon release or if the gripped object collides with another, continuing until contact ends.

\subsubsection{Kinesthetic Teaching with Force Feedback (KTF)}
% ktf
Building on the initial KT interface, force feedback is introduced by transmitting the forces experienced by the virtual robot’s joints to the physical robot’s joints in real time. In the simulation, external forces on the virtual robot’s joint torque sensors are scaled and applied to corresponding joints of the physical robot. This process enhances users' perception of physical forces while maintaining a safe environment, making manipulation more intuitive and responsive.
% Based on the initial KT interface, explicit force feedback is implemented by transmitting the forces experienced by the virtual robot’s joints to the physical robot’s joints in real time. In the simulation, external forces on the virtual robot’s joint torque sensors are monitored, smoothed, scaled, and applied to the corresponding joints of the physical robot. This enhances the operator’s perception of physical forces while ensuring a safe environment, making the manipulation process more intuitive and responsive.

\subsection{User Study}
A user study was designed to compare the force feedback interface with the original one, assessing their impact on data collection efficiency, effectiveness, and user experience across various tasks and identifying significant differences.
% To assess the influence of various force-feedback interfaces on data collection efficiency, effectiveness, and user experience, the user study was designed to evaluate the interface with force-feedback functionality in comparison to the original interface and to compare the different impacts of these interfaces across various demonstration tasks, while exploring statistically significant differences between them.
\subsubsection{Task Design}

\begin{figure*}[h]
    \centering
    % First row (VR Tasks) without subcaptions
    \begin{minipage}{\textwidth}
        \centering
        \includegraphics[width=0.24\textwidth]{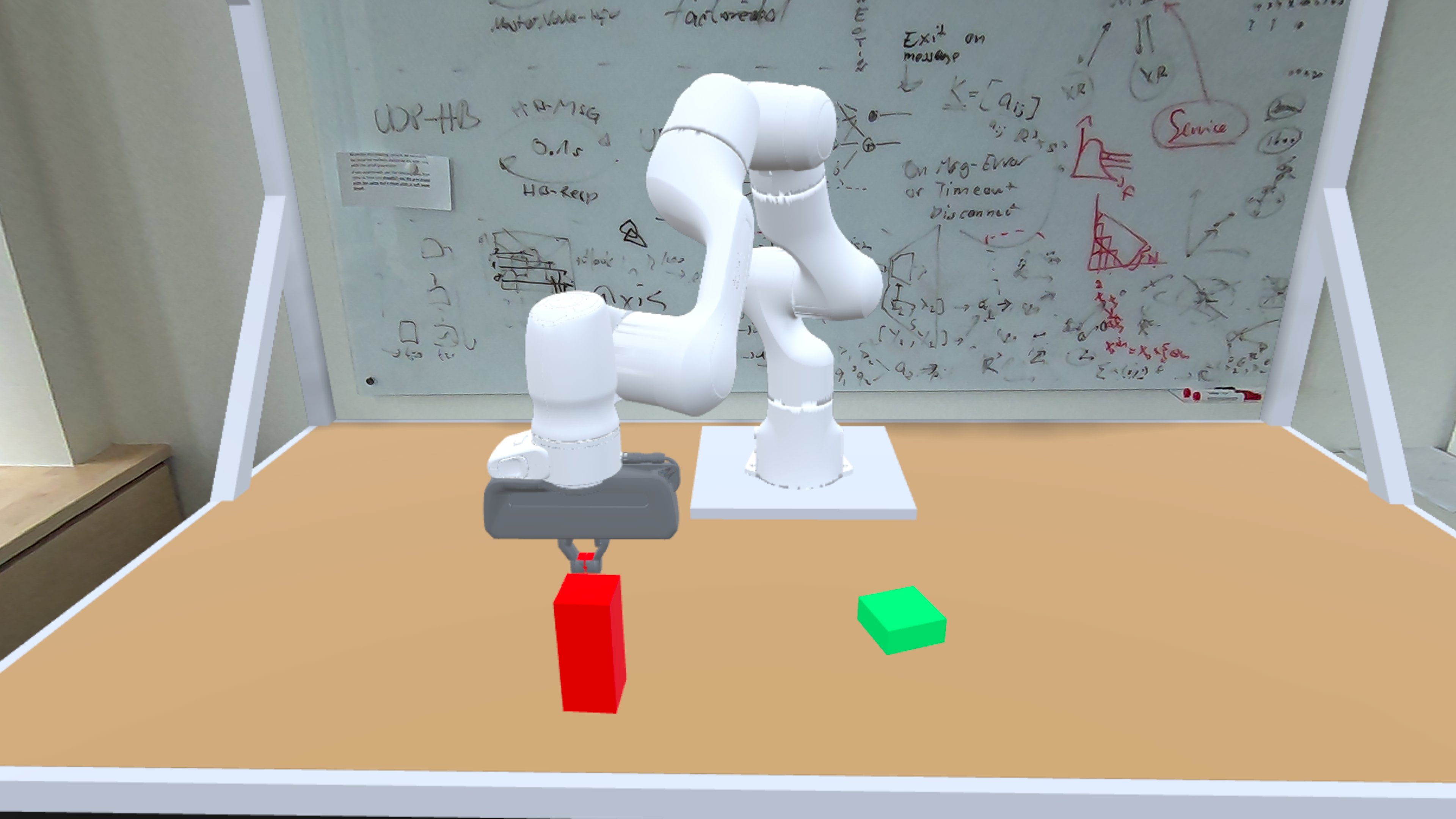}%
        \hspace{0.5mm}%
        \includegraphics[width=0.24\textwidth]{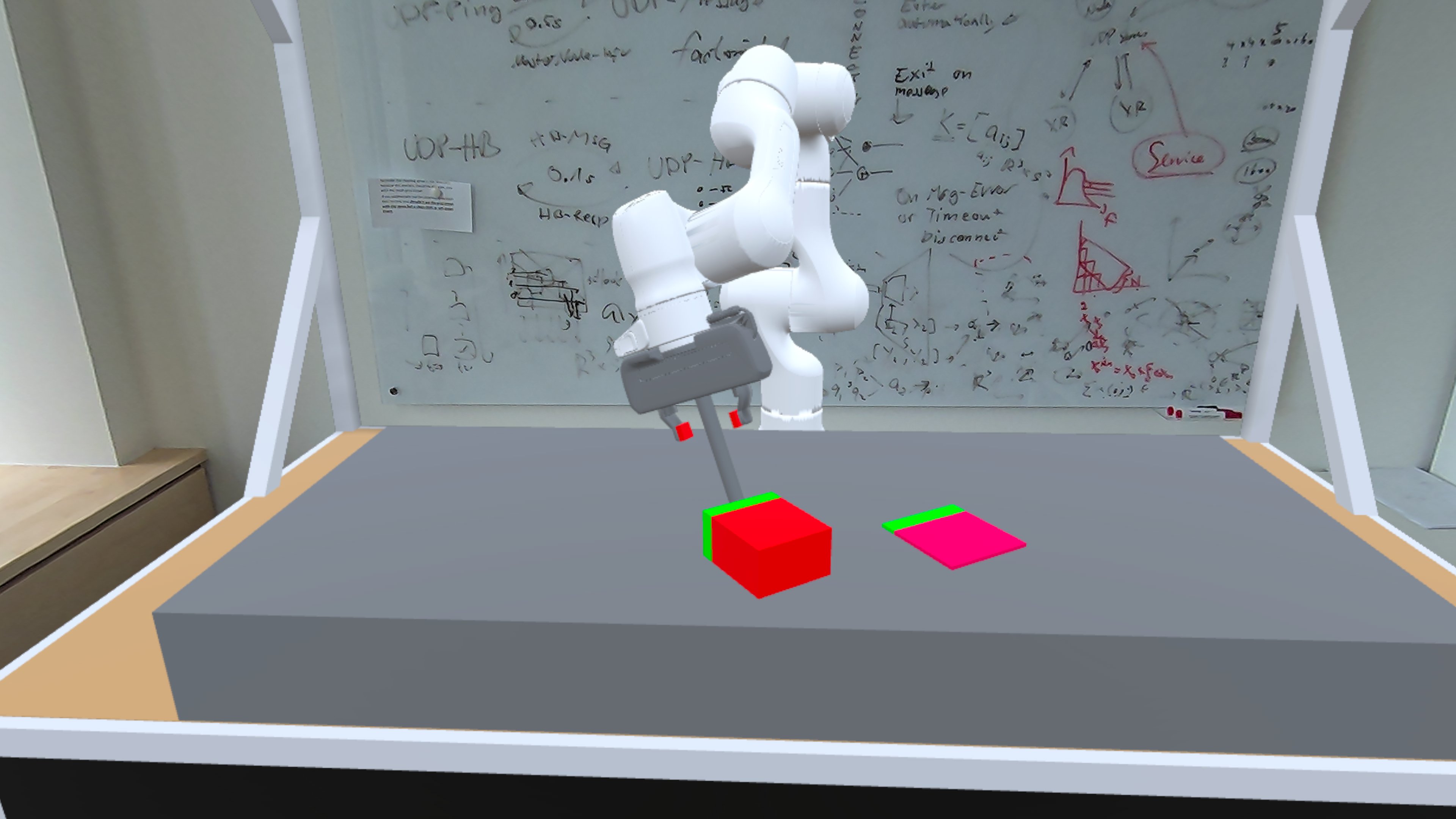}%
        \hspace{0.5mm}%
        \includegraphics[width=0.24\textwidth]{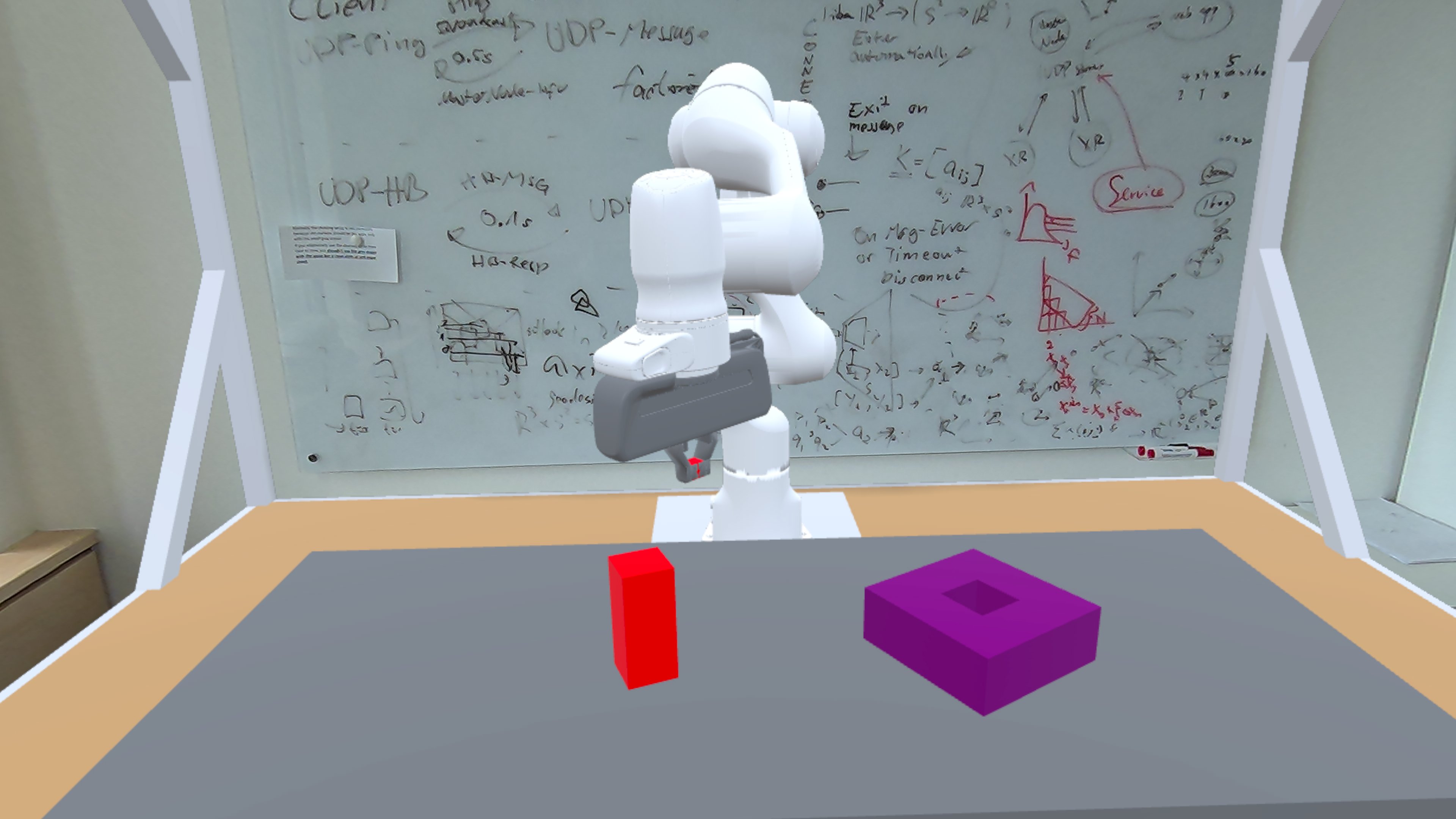}%
        \hspace{0.5mm}%
        \includegraphics[width=0.24\textwidth]{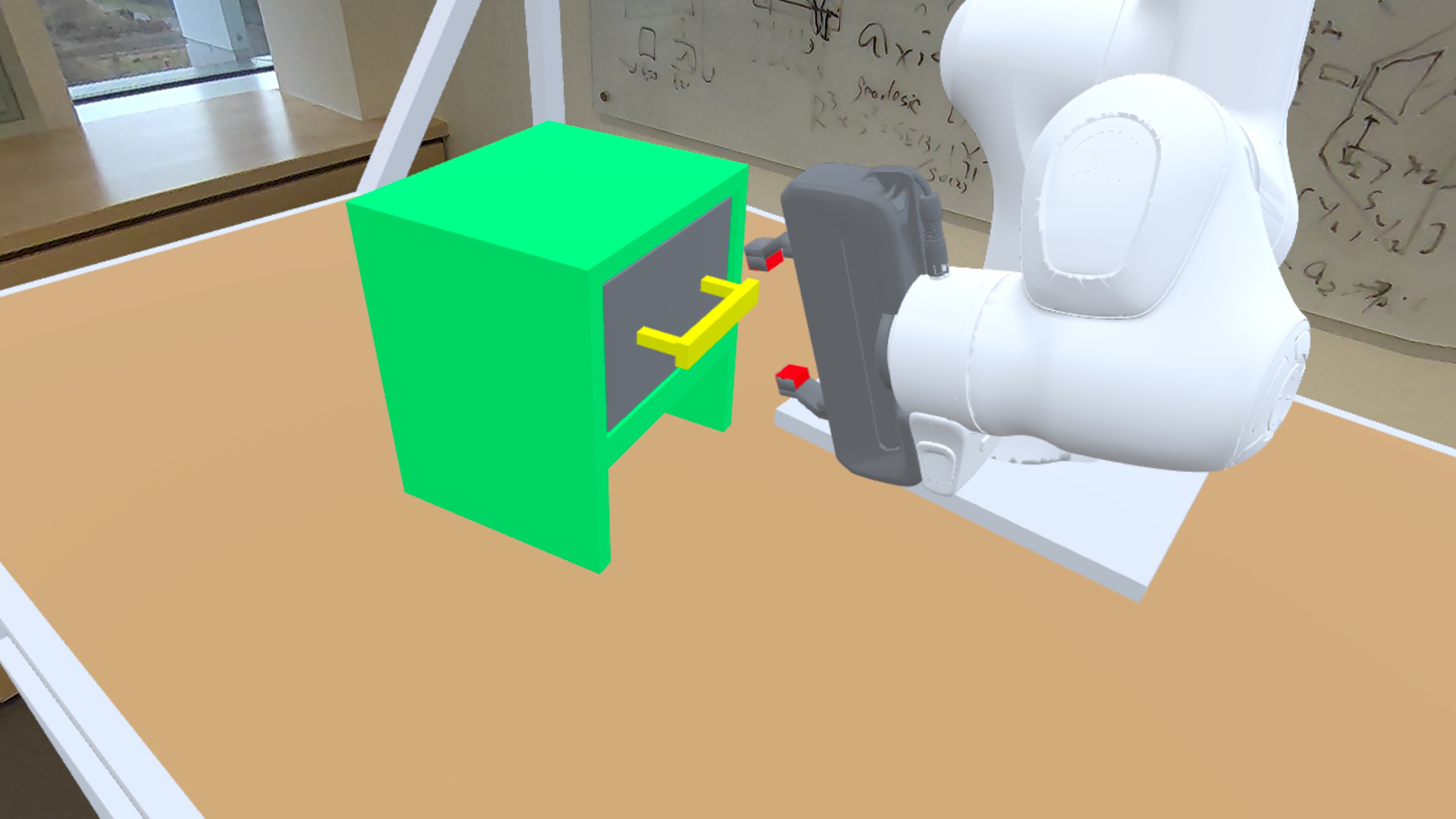}
    \end{minipage}

    \vspace{0.5mm}%

    % Second row (Standard Tasks) with subcaptions
    \begin{minipage}{\textwidth}
        \centering
        \subfloat[Pick Task]{\includegraphics[width=0.24\textwidth]{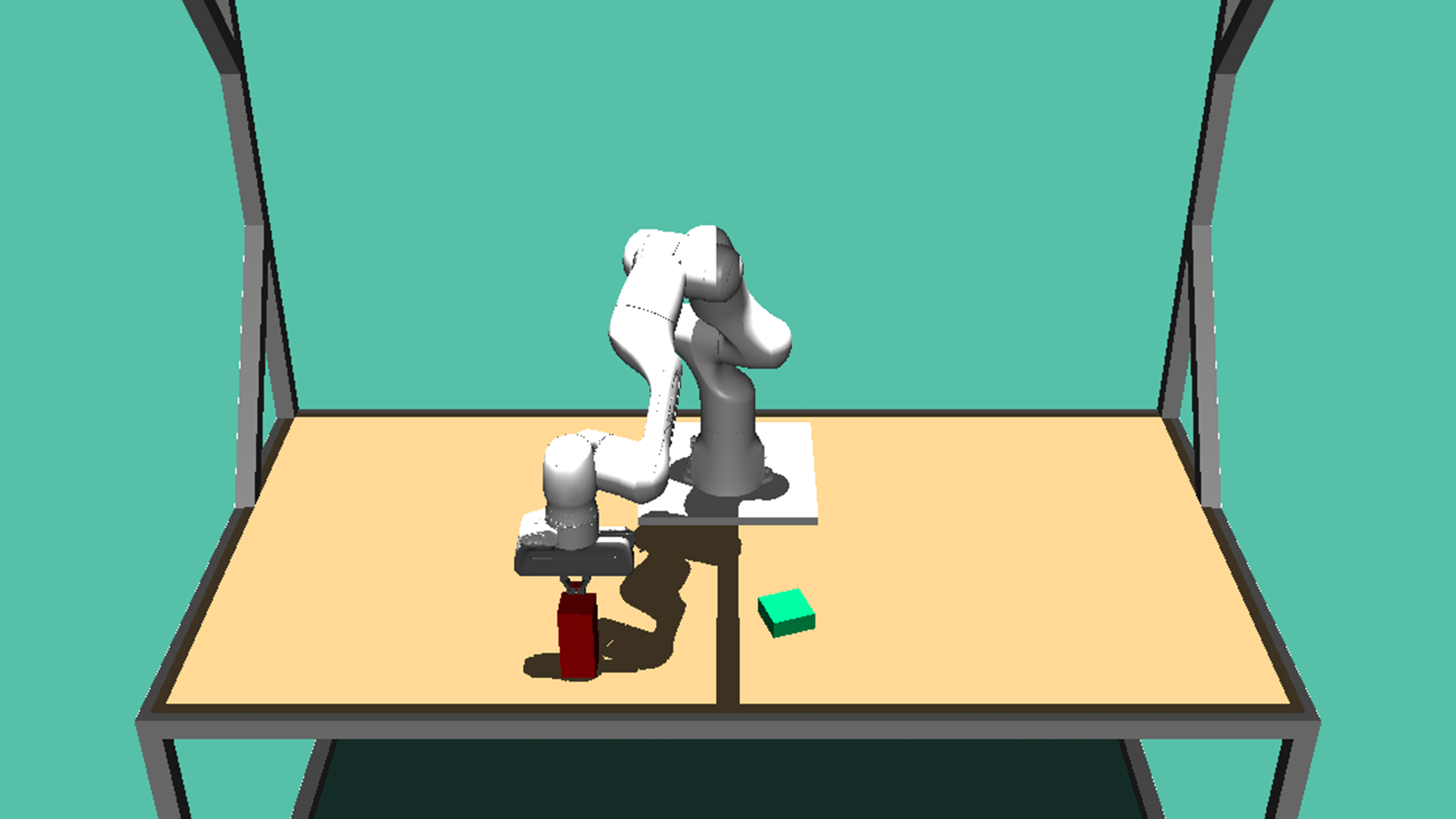}}%
        \hspace{0.5mm}%
        \subfloat[Push Task]{\includegraphics[width=0.24\textwidth]{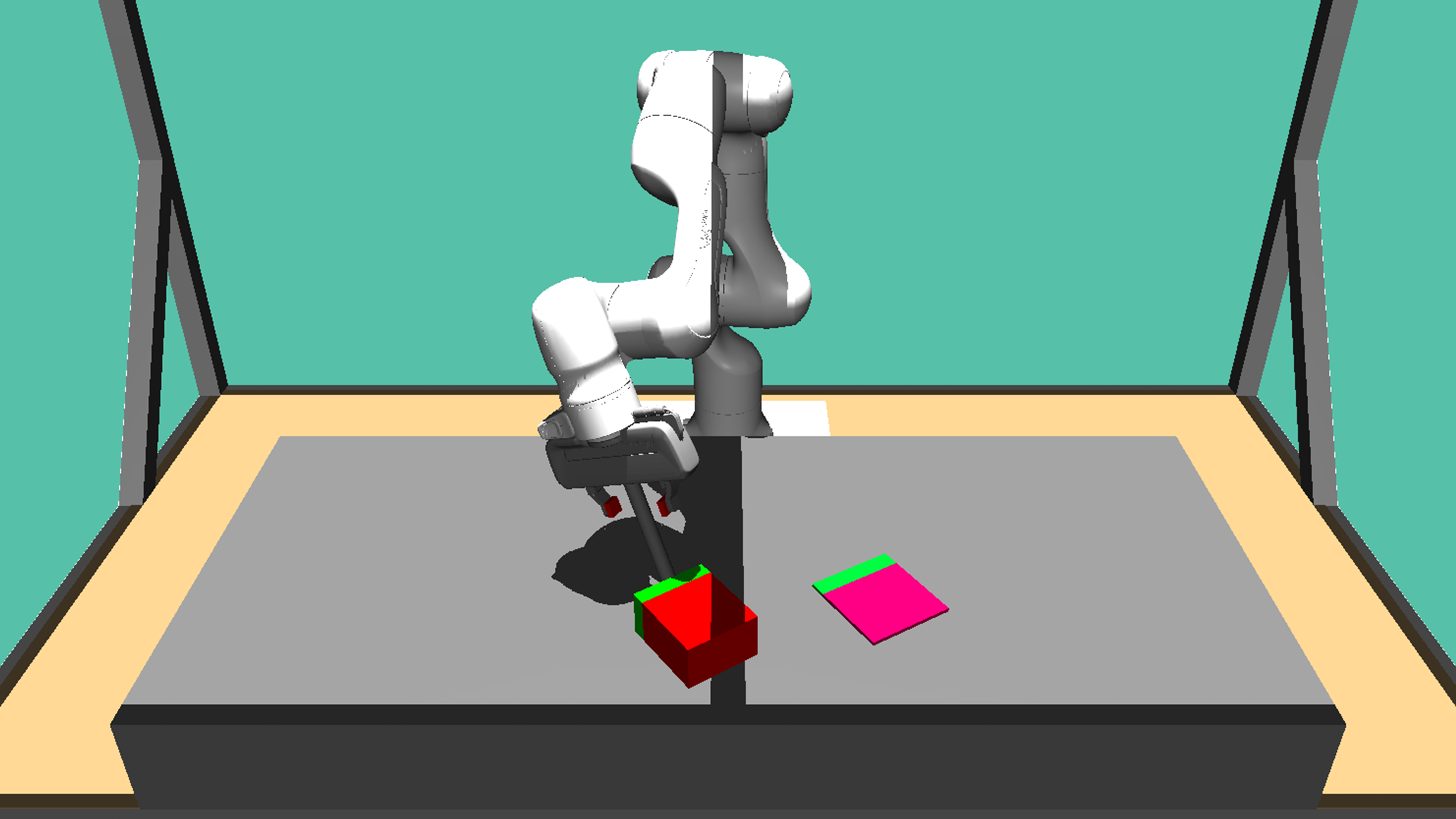}}%
        \hspace{0.5mm}%
        \subfloat[Assemble Task]{\includegraphics[width=0.24\textwidth]{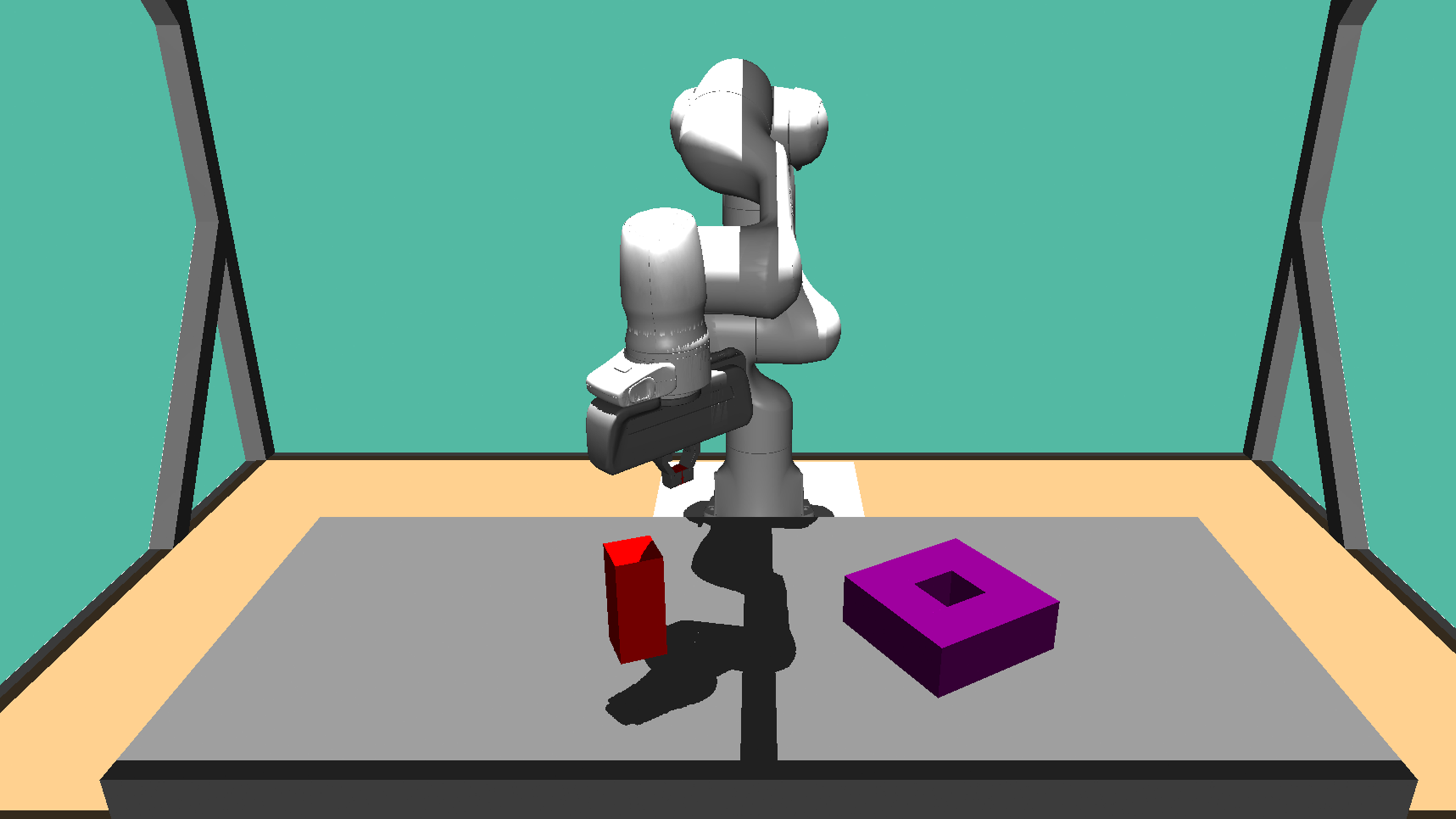}}%
        \hspace{0.5mm}%
        \subfloat[Open Task]{\includegraphics[width=0.24\textwidth]{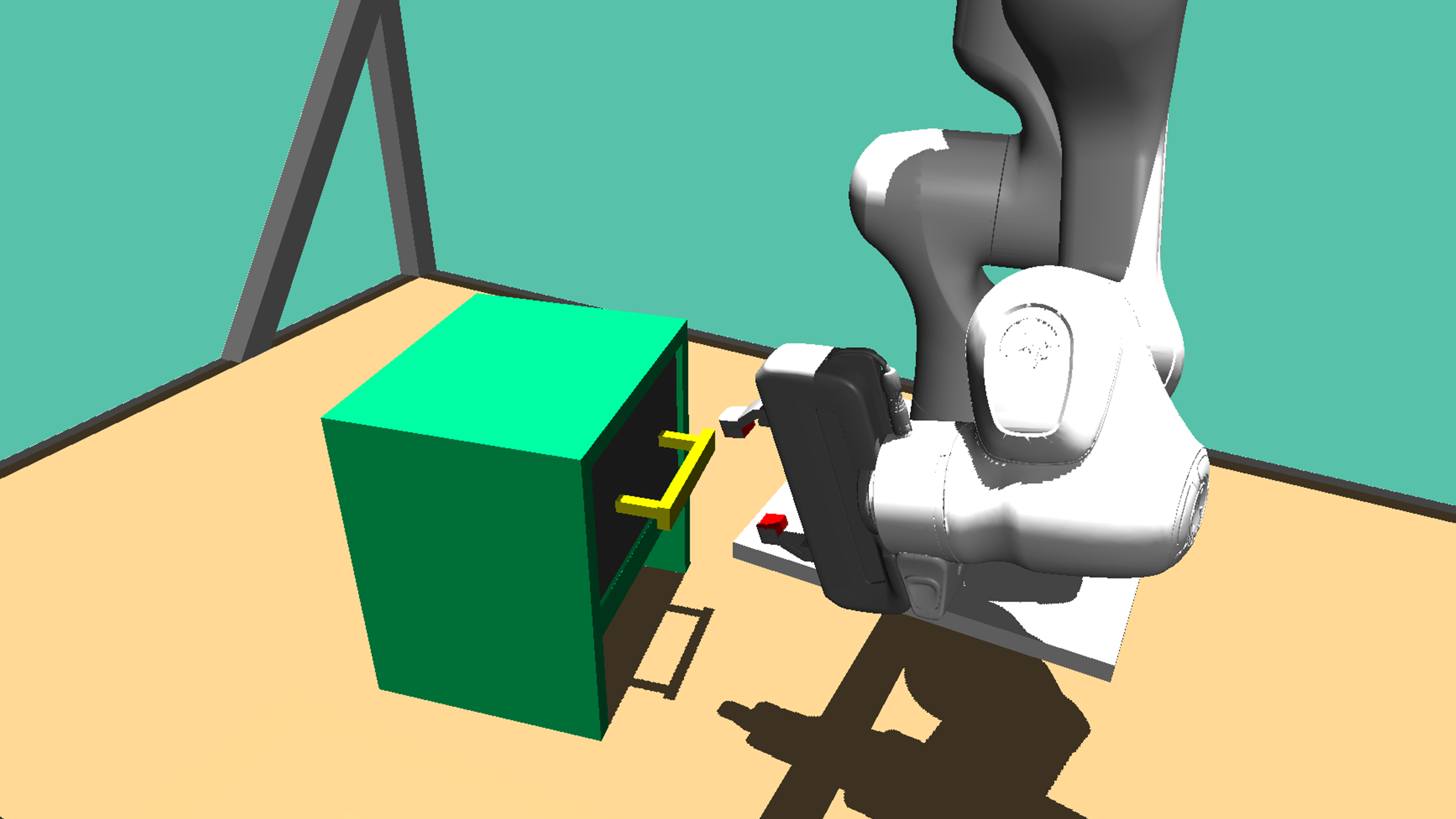}}
    \end{minipage}

    \caption{
The four tasks used in the user study to evaluate the performance of each interface. The top row shows images captured from VR headsets, while the bottom row presents the corresponding tasks in simulation.
    }
    \label{fig:task_comparison}
\end{figure*}

As shown in Fig. \ref{fig:task_comparison}, four demonstration tasks were chosen for the user study: \textit{Push Cube}, \textit{Pick and Place}, \textit{Assemble Peg}, and \textit{Open Drawer}.
% These tasks, ranging from basic to complex, are common in robot manipulation tasks and involve both simple and precise manipulations.
% Four demonstration tasks were selected for the user study: \textit{Push Cube}, \textit{Pick and Place}, \textit{Assemble Peg} and \textit{Open Drawer}. These are common tasks in LfD, including both basic and precise manipulations, and operations from simple to complex.

\textbf{Push Cube.} Participants push a cube into a target area,
% aligning a colored face.
The task is considered complete when the cube is within the target area and the colored face is correctly aligned.
% Time limit: 60 seconds.

\textbf{Pick and Place.} Participants pick up a box and place it on a target platform without overlapping the edges. This task and Push Cube are two common manipulation tasks designed to evaluate the fundamental manipulation skills of a robot control interface.
% Time limit: 60 seconds.

\textbf{Assemble Peg.} Participants insert a 60 mm × 60 mm square peg into a 64 mm × 64 mm square hole. The task is complete when the peg is securely inserted with no misalignment. If the peg is dropped, the task fails. This task requires highly accurate movement from operators, which is used to assess the responsiveness and precision of robot control interfaces.
% Time limit: 60 seconds.

\textbf{Open Drawer.} Participants align and grip the drawer handle, pulling it out smoothly. The task is complete when the drawer is open but not fully extended. This task, involving compound movements, evaluates the comprehensive manipulation capabilities of the robot control interface.

These tasks, ranging from basic to complex, are common in robot manipulation tasks and involve both simple and precise manipulations.

\subsubsection{Objective Metrics Design}

% \textbf{Objective Metrics} :
% \textit{Success rate} and \textit{completion time} were utilized as objective metrics to evaluate the interfaces.  
% \textit{Success rate} indicates whether the demonstration task was completed according to the specified requirements using the designated interface, recorded as 0 (failure) or 1 (success).
% \textit{Completion time} represents the duration participants took to complete the demonstration task. If the participant failed or exceeded the limited time for a task, the maximum allowed time was recorded instead. The time limits for each task were determined based on a pilot user study and previous research. This approach ensured that participants had sufficient time to complete the demonstration task without extending the overall duration of the user study excessively. 

To comprehensively assess each interface objectively,
% both objective and subjective metrics were used in this study.
\textit{Task success} and \textit{Task completion time} were employed as objective metrics.
The \textit{task success} indicates whether a task was completed as required, recorded as 0 (failure) or 1 (success). \textit{Task completion time} represents the duration taken to finish a task,
with the maximum allowed time recorded if the participant failed or exceeded the time limit.
If the task was not completed within the time limit, the task success was recorded as 0.
Task time limits were determined based on a pilot study,
% and prior research,
ensuring sufficient time for task completion without extending the overall duration of the user study.
In this study, the time limit for all tasks is set to 60 seconds.

% \subsubsection{Questionnaire Design}
\subsubsection{Subjective Metrics Design}
In addition to the objective metrics, it is also important to asses how the user experiences each interface.
Hence, for subjective metrics, 
two types of questionnaires are used to test hypotheses: overall subjective metrics and task-wise subjective metrics.
The overall subjective questionnaire evaluates the overall performance of the four interfaces by \textit{User Experience Questionnaire Short Version (UEQ-S)} \cite{schrepp2017design}.
UEQ-S measures interface quality across two dimensions: pragmatic quality and hedonic quality, each with four items, with the scale from -3 to +3.
The task-wise subjective questionnaire was designed to evaluate user experience for each individual task. Four task-specific metrics—accuracy, stability, efficiency, and usability—were assessed using a 5-point Likert scale to capture participants’ perceptions of the interfaces.
Accuracy measures the effectiveness of the interface and the user's understanding of the system. Stability evaluates the reliability of the interface, which directly influences user trust. Efficiency quantifies the time and effort required to complete tasks, serving as an indicator of cognitive load. Usability reflects participants’ overall perception of the interface’s ease of use. This structured evaluation provides insights into the impact of force feedback on different interfaces in a task-specific context.

\subsubsection{User Study Progress}
To minimize the impact of participants' proficiency on the study,
they are given a warm-up session before using each interface to familiarize themselves with its operations.
During the user study, each task allowed up to five attempts to account for proficiency and self-efficacy effects \cite{locke1997self}.
Participants could choose when they were satisfied with their performance, while the researcher recorded task completion time and task success.

After completing each task, participants were required to fill out task-wise subjective metric assessments to evaluate the interface's performance for the given task. Upon completing demonstrations for all four tasks within an interface, they completed the UEQ-S questionnaire for an overall evaluation.

To minimize biases from task sequencing and fatigue, a randomized design \cite{wickens2004design} was used.
Both the order of interfaces and the sequence of tasks were randomized to mitigate systematic effects.
% To minimize the influence of participants' proficiency on demonstration outcomes and subjective perceptions, a familiarization task was introduced before each interface. This task allowed participants to practice and understand the interface's operational methods before processing the demonstration tasks. As the interface was not designed for first-time users, each task permitted up to five attempts to account for proficiency and self-efficacy effects \cite{locke1997self}. Participants could decide when they were satisfied with their performance, while the researcher recorded the completion time. 

% % Following each demonstration, participants completed a subjective evaluation scale assessing the interface's performance within the given task. After completing four task demonstrations with an interface, they filled out the UEQ-S questionnaire to assess their overall experience.

% To counteract learning effects, fatigue, and potential biases due to task sequencing, a randomized design \cite{wickens2004design} was implemented. Both the order of the four interfaces and the sequence of demonstration tasks were randomized to minimize systematic effects.

\subsubsection{Participants}
The user study involved 31 participants, aged 20 to 35, including three females and 28 males.
% To standardize subjective factors,
All participants were students or university staff.
% with experience in using or developing robots, as the interfaces were designed for professional users.
% Four participants reported prior experience with Meta Quest or other VR devices.
Each participant interacted with all four interfaces in a randomized task order, resulting in 496 valid human demonstrations.

\section{RESULTS AND DISCUSSIONS}

This section presents the results of the user study, including both objective and subjective metrics. This is followed by a discussion on whether the findings align with the three hypotheses proposed in Section \ref{sec:introduction}.
Since the dependent variables (objective metrics and task-wise subjective metrics), as well as the factors (task types and the presence or absence of force feedback), are not homogeneous, the Mann-Whitney U test \cite{mcknight2010mann} is applied to determine whether each factor leads to significant differences in the metrics.

% The study involves four different interfaces, making each interface a categorical factor of the independent variable. First, by analyzing the averages of the results, a preliminary understanding of the performance differences across interfaces can be gained. Since the independent variable, interface with and without force feedback, is categorical, and the dependent variables (success rate, completion time) are ordinal or continuous but not homogeneous, the Mann-Whitney U test \cite{mcknight2010mann} was applied to examine whether the presence or absence of force feedback results in significant differences in the objective metrics. 
% Additionally, for the different demonstration tasks, the results were analyzed independently for each task.
\subsection{Objective Metrics}

\subsubsection{Success Rate}
% \input{Tables/result/objective_metrics/difference}
% \input{Figure-success}
% For success rate, the collected data are binary (0 or 1) and not normally distributed.
% Descriptive Statistics - Mean
As shown in Tab. \ref{table:success_rate},
the average success rate of each task varies on different interfaces. 
The result shows that the introduction of force feedback led to an improvement in the MC interface, which supports \textbf{H1}.
The effect of force feedback was particularly notable in the Assemble Task, which requires highly precise manipulation.
Especially, MCV has a success rate of $87\%$ from Assemble Task.
It indicates that the introduction of force feecback improves performance nearly $30\%$.
By Mann-Whitney U test,
there is only one significant difference in success rate was found in the Assemble Task between MC and MCV ($p < 0.05$).
% With force feedback, the success rate of this task improved considerably ,
% while no statistically significant differences were observed in other cases.
From all the success rate result, the KT and KTF interfaces consistently maintained higher success rates.
% As shown in Table\ref{table:difference_object}, with the p-value less than 0.05, it can be observed that a significant difference in success rate was found in the Assemble Task only under the MC interface when force feedback was introduced. With force feedback, the task success rate was considerably improved in the Assemble Task, in all other cases, no statistically significant differences were detected.
\begin{table}[H]
\centering
\caption{Mean of success rate for different tasks with and without force feedback.}
\begin{tabular}{lcccc}
\toprule
\textbf{Interface} & \textbf{Push Task} & \textbf{Pick Task} & \textbf{Assemble Task} & \textbf{Open Task} \\ \midrule
MC   & 94\%    & 97\%    & 58\%    & 93.55\% \\
MCV  & 100\%   & 97\%    & 87\%    & 100\% \\
KT   & 100\%   & 100\%   & 97\%    & 100\% \\
KTF  & 100\%   & 100\%   & 100\%   & 100\% \\
\bottomrule
\end{tabular}

\label{table:success_rate}
\end{table}

\subsubsection{Completion Time}

\begin{figure}[h]
    \centering
    \includegraphics[width=0.48\textwidth]{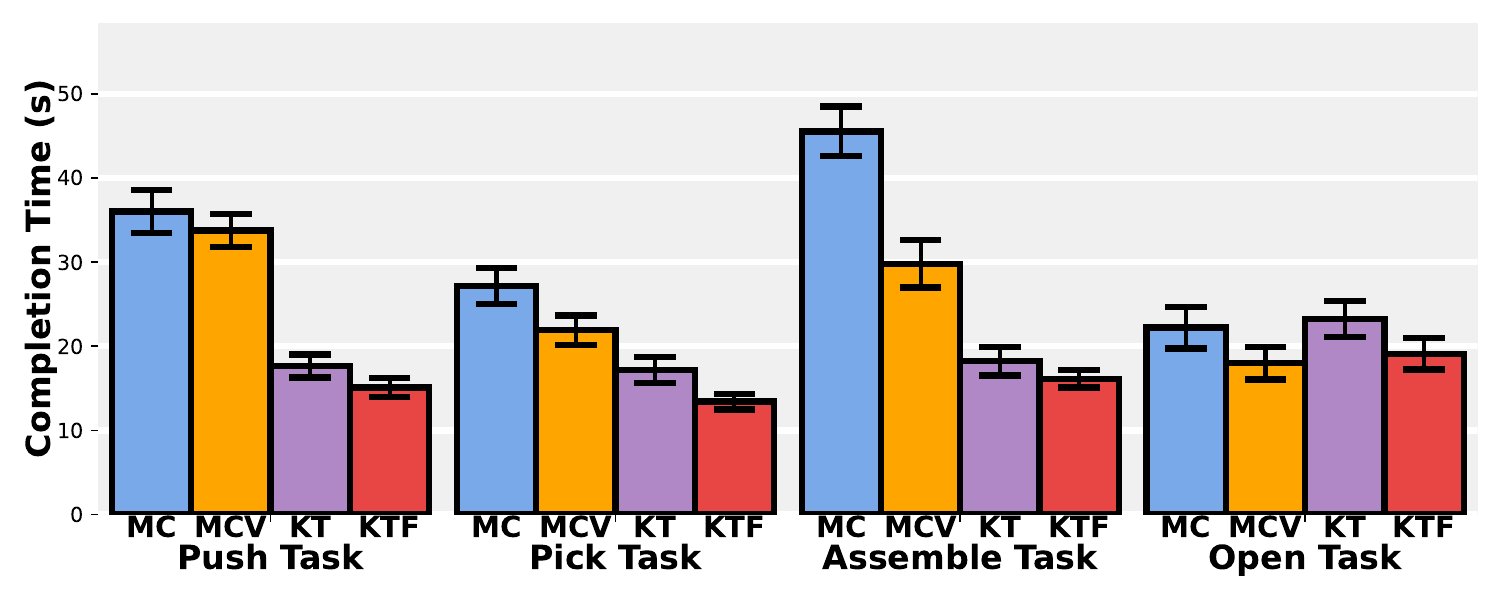}
    \caption{Mean and standard error of completion time for different tasks and interfaces. 
    The bars represent the mean completion time for each interface (MC, MCV, KT, KTF) across 
    different tasks (Push, Pick, Assemble and Open).
    The thick black outlines indicate the standard error, 
    showing the variability in task completion times.}
    \label{fig:mean_completion_time}
\end{figure}

% For completion time, which is a continuous variable, the distribution is approximately normal, but the variances are not homogeneous. 
The completion times with means and standard error are listed in Fig. \ref{fig:mean_completion_time}.
Completion times were shorter for interfaces with force feedback compared to those without, supporting \textbf{H1}.
% The standard error of MVC and KTF are relatively smaller compared to initial interfaces across all tasks, indicating its consistently stable performance during data collection.
% \input{Tables/result/objective_metrics/time}
Similar to the results of success rate,
a highly significant difference in completion time was observed only in the Assemble Task under the MC interface when force feedback was introduced, with $ p < 0.01$.
With force feedback, the completion time was significantly reduced in the Assemble Task.
In all other cases, no statistically significant differences were detected, supporting \textbf{H3}.

\subsection{Subjective Metrics}

\subsubsection{Overall Subjective Metrics}
By using the UEQ-S, overall subjective assessments of the interfaces were obtained.
Following the UEQ-S guidelines, mean values for each item were computed to assess participants’ perceptions of pragmatic quality, hedonic quality, and overall subjective evaluation. 
\begin{table}[ht]
\centering
\caption{Mean and standard deviation of results for UEQ-S questionnaire in different interfaces}
\begin{tabular}{lccc}
\toprule
\textbf{Interface} & \textbf{Pragmatic Quality} & \textbf{Hedonic Quality} & \textbf{Overall} \\ \midrule
MC  & $0.90 \pm 0.88$  & $1.07 \pm 0.98$  & $0.99 \pm 0.73$  \\ 
MCV & $1.11 \pm 0.90$  & $1.44 \pm 0.88$  & $1.27 \pm 0.73$  \\ 
KT  & $2.00 \pm 0.61$  & $1.34 \pm 0.96$  & $1.67 \pm 0.69$  \\ 
KTF & $\mathbf{2.13 \pm 0.66}$  & $\mathbf{1.60 \pm 0.92}$  & $\mathbf{1.86 \pm 0.69}$  \\ 
\bottomrule
\end{tabular}
\label{table:mean_ueqs}
\end{table}

The results from Tab. \ref{table:mean_ueqs} indicate that the force feedback led to a notable increase in scores for pragmatic quality, hedonic quality, and overall evaluation, supporting \textbf{H1}.
Additionally, the standard deviation in hedonic quality and overall evaluation scores was reduced, suggesting that force feedback enhances participants’ subjective experience.
Compared to the KT interface, the MC interface used to control a virtual robot exhibited a more pronounced improvement in subjective ratings after the introduction of force feedback, supporting \textbf{H2}.

\subsubsection{Task-wise Subjective Metrics}
\begin{figure}[h]
    \centering
    \vspace{-1mm}
    \begin{subfigure}[b]{0.48\textwidth}
        \centering
        \includegraphics[width=\textwidth]{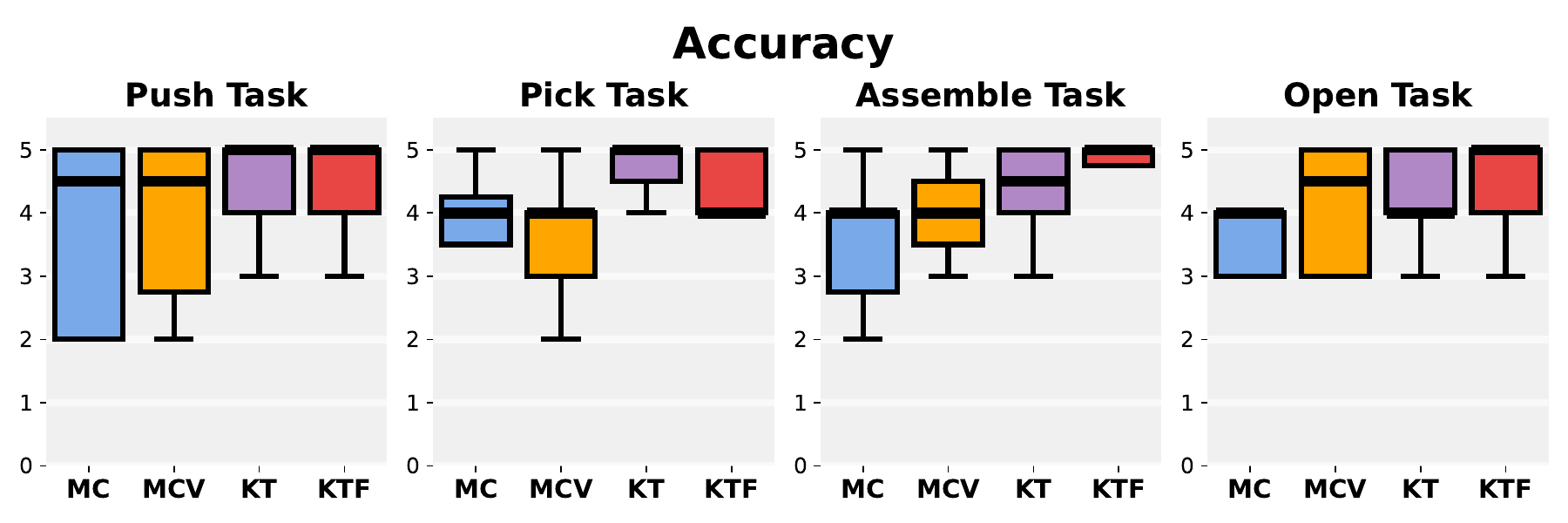}
        % \caption{Accuracy}
    \end{subfigure}%
    \vspace{-1mm} % Adjust spacing if needed
    \begin{subfigure}[b]{0.48\textwidth}
        \centering
        \includegraphics[width=\textwidth]{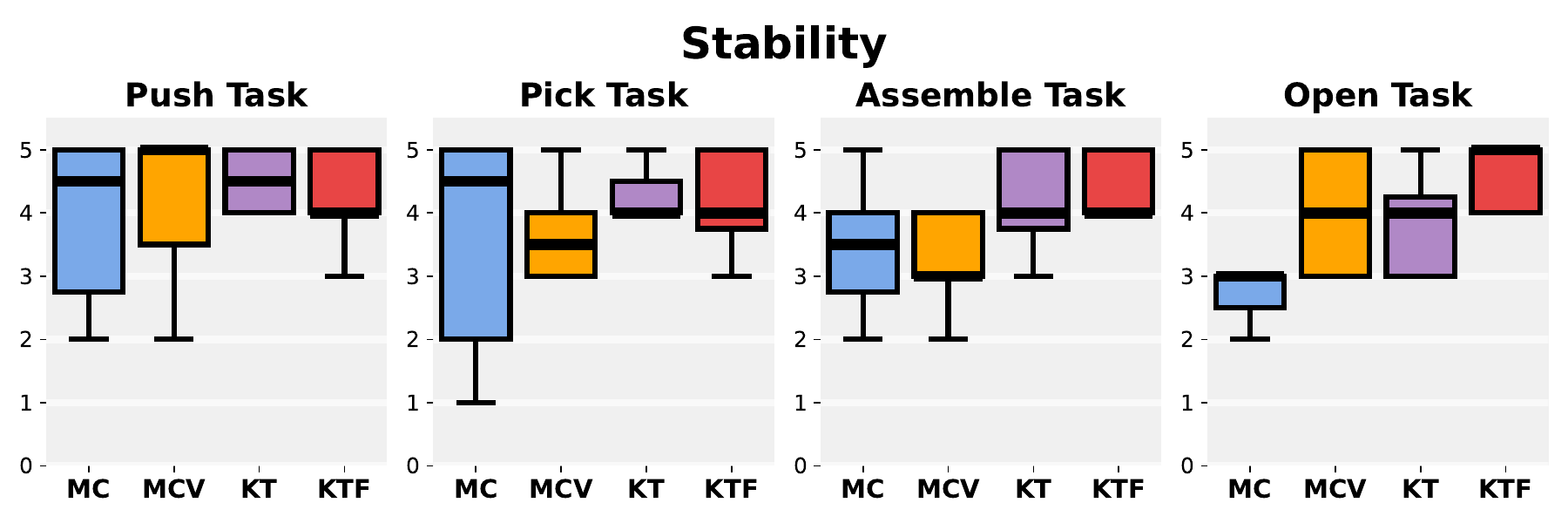}
        % \caption{Stability}
    \end{subfigure}%
    \vspace{-1mm} % Adjust spacing if needed
    \begin{subfigure}[b]{0.48\textwidth}
        \centering
        \includegraphics[width=\textwidth]{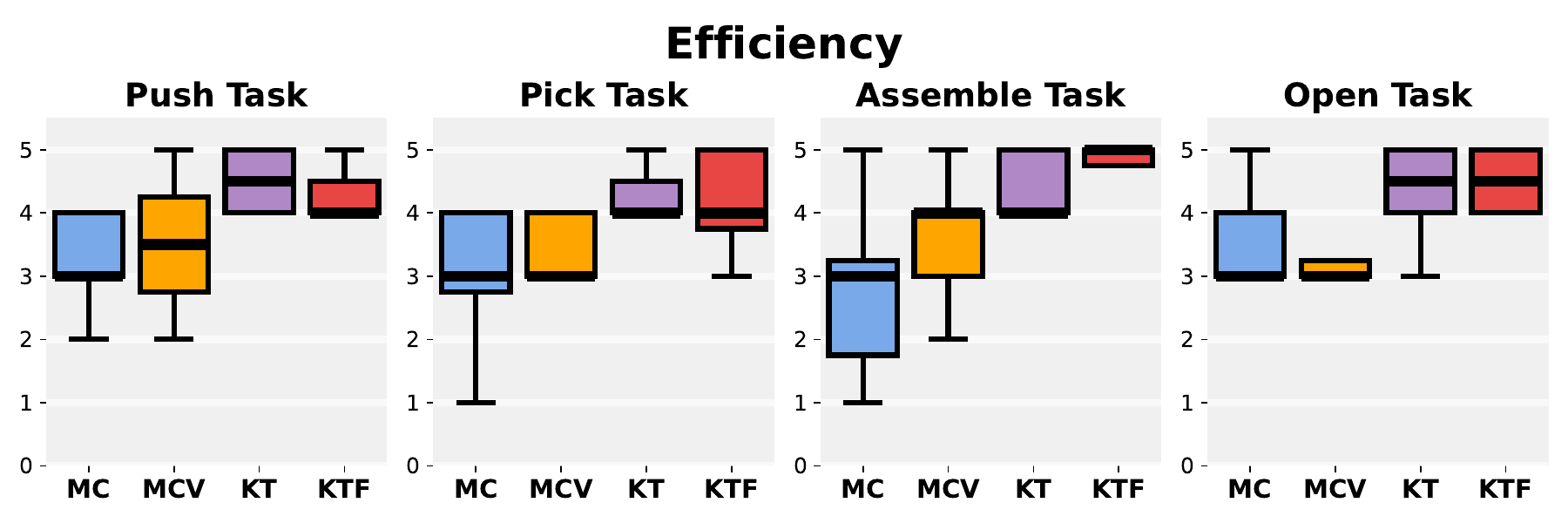}
        % \caption{Efficiency}
    \end{subfigure}%
    \vspace{-1mm} % Adjust spacing if needed
    \begin{subfigure}[b]{0.48\textwidth}
        \centering
        \includegraphics[width=\textwidth]{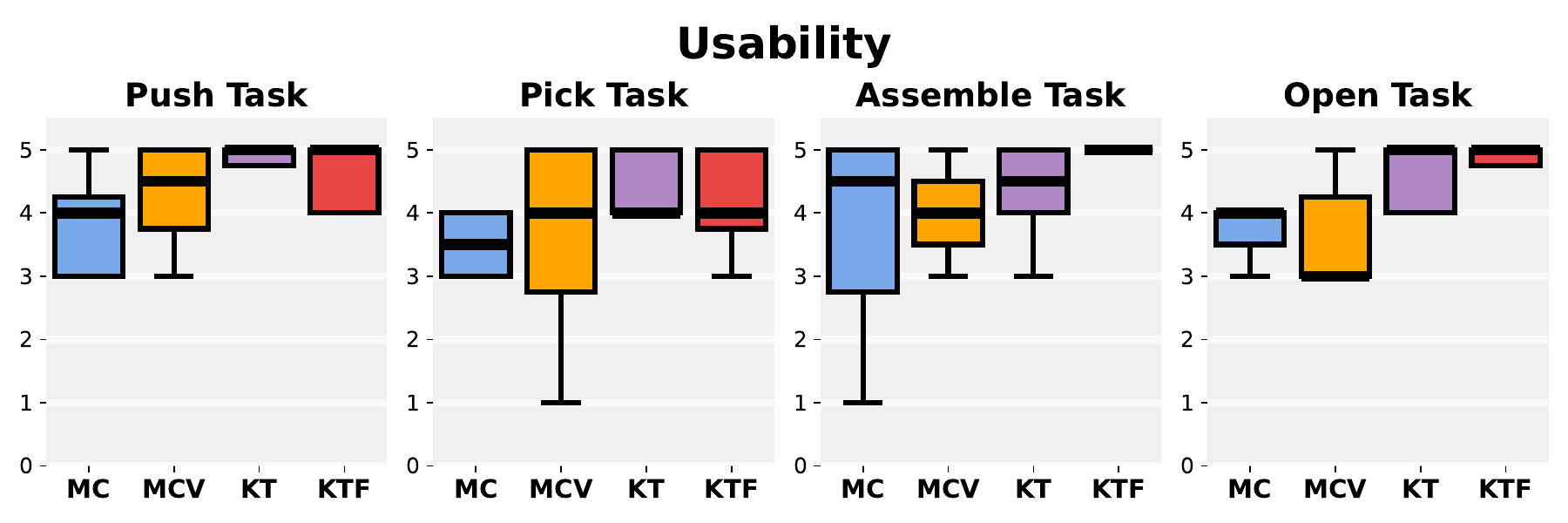}
        % \caption{Usability}
    \end{subfigure}

    \caption{Boxplots for different subjective metrics across tasks and interfaces.}
    \label{fig:subjective_metrics}
\end{figure}

The subjective evaluations of each interface and task are presented in Fig. \ref{fig:subjective_metrics}.
Likewise, the introduction of force feedback led to an overall improvement in these subjective metrics.

\begin{table}[h]
\centering
\caption{P-values for differences in subjective metrics with and without force feedback (*: $p < 0.05$ indicates significant difference, $-$: $p > 0.5$ indicates no significant difference)}
\resizebox{0.5\textwidth}{!}{
\begin{tabular}{l|lcccc}
\toprule
\textbf{Metric} & \textbf{Interface} & \textbf{Push Task} & \textbf{Pick Task} & \textbf{Assemble Task} & \textbf{Open Task} \\ 
\midrule
\multirow{2}{*}{\textbf{Accuracy}} 
                & MC-MCV  & 0.352 & 0.272 & 0.019* & 0.447 \\ \cline{2-6} 
                & KT-KTF  & - & - & 0.141 & - \\ \hline
\multirow{2}{*}{\textbf{Stability}} 
                & MC-MCV  & 0.253 & 0.298 & 0.061 & 0.427 \\ \cline{2-6} 
                & KT-KTF  & 0.331 & - & 0.248 & 0.301 \\ \hline
\multirow{2}{*}{\textbf{Efficiency}} 
                & MC-MCV  & - & 0.049* & 0.019* & - \\ \cline{2-6} 
                & KT-KTF  & - & - & 0.174 & 0.433 \\ \hline
\multirow{2}{*}{\textbf{Usability}} 
                & MC-MCV  & 0.426 & 0.045* & 0.014* & 0.434 \\ \cline{2-6} 
                & KT-KTF  & - & - & - & - \\ 
\bottomrule
\end{tabular}
}
\label{table:difference_subject}
\end{table}

% Similar to the analysis of objective metrics, 
% The Mann-Whitney U test \cite{mcknight2010mann} was used to assess whether the introduction of force feedback had a significant impact on subjective evaluations within the same interface. 
As shown in Tab. \ref{table:difference_subject}, for accuracy, a statistically significant difference ($p < 0.05$) due to force feedback was observed only in the Assemble Task with the MC interface.
For efficiency,
significant differences ($p < 0.05$) were found in the Pick Task and Assemble Task when using the MC interface.
Similarly, force feedback led to significant improvements ($p < 0.05$) in the Pick Task and Assemble Task with the MC interface in the usability metrics.
In other cases, particularly with the KT interface,
although subjective scores increase when introducing force feedback,
the differences were not statistically significant.

\subsection{Discussion}
This study explored the impact of force feedback on data collection efficiency, effectiveness, and user experience across various tasks.
The results show that force feedback generally improved performance and user experience, with varying effects depending on task complexity and interface type.
For tasks like Assemble Peg, requiring precise repositioning, force feedback in the MC interface significantly improved success rate and completion time.
In contrast, other tasks showed improving trends, but differences were not statistically significant.
% This study explored the impact of introducing force feedback into different interfaces on data collection efficiency, effectiveness, and user experience across various demonstration tasks.
% The results indicate that force feedback generally enhances performance and user experience, with varying degrees of impact depending on task complexity and interface type. Tasks requiring precise repositioning, such as the Assemble Task, exhibited significant improvements in success rate and completion time when force feedback was applied to the MC interface. In contrast, other tasks, including those performed using the KT interface, showed an improving trend, though the differences were not statistically significant. 

Subjective evaluations further support following findings.
The results from the UEQ-S indicate that the introduction of force feedback led to notable increases in scores for pragmatic quality, hedonic quality, and overall evaluation.
Force feedback in the MC interface significantly improved perceived accuracy, efficiency, and usability in the Assemble Task.
Additionally, significant improvements in efficiency and usability were observed in the Pick Task.
% In contrast, for other tasks, while force feedback resulted in minor improvements in subjective ratings,
% these differences did not reach statistical significance.
In contrast, other tasks showed no statistically significant improvements in subjective ratings.
Under the KT interface, force feedback led to a slight increase in subjective ratings,
but the effect was less pronounced compared to the MC interface.

From the perspective of the impact of different types of force feedback,
introducing force feedback to the MC interface significantly influenced some demonstration tasks.
In contrast, the KT interface only showed a positive trend without reaching statistical significance.
This suggests that force feedback provides a more pronounced experiential difference in the MC interface, which benefited more from it than the KT interface.
When manipulating the MC interface, the operator is at a certain distance from the virtual robot and the manipulated objects.
For tasks that require object repositioning, purely visual cues may be insufficient, especially when visual occlusion occurs.
Force feedback can provide easily perceptible haptic cues that complement the visual information.
In contrast, when using the KT interface, the virtual and real robots need to overlap, and the operator stands closer to the robot and can observe it more directly, which makes visual cues sufficient.

Regarding the impact of force feedback across different tasks, it is evident that task complexity influences sensitivity to force feedback.
Complex tasks, such as the Assemble Task, benefit significantly from force feedback, while simpler tasks, like the Open Task, exhibit only minor improvement trends.

In the Open task, participants demonstrated low sensitivity to interface changes.
Opening the drawer typically relies on visual cues—gripping the handle and pulling it to a predefined position—so the force feedback provides minimal benefit, resulting in only slight, non-significant improvements.
Similarly, in the Push Task, a relatively simple operation, solely using visual feedback suffices, and force feedback primarily serves to confirm contact between the end-effector and the object, resulting in only slight, non-significant improvements. In the Pick Task, where visual occlusion or distant target object may be involved, force feedback significantly enhanced subjective ratings of efficiency and usability, though other effects remained non-significant. For the most complex task, the Assemble Task, which requires high-precision actions, force feedback proved particularly beneficial by aiding the operator in confirming positional cues and avoiding errors. Thus, in such complex, precision-demanding tasks, force feedback significantly enhances efficiency, effectiveness, and user experience.

These findings align with the proposed hypotheses:
\textbf{H1:} Force feedback generally improved efficiency, effectiveness, and user experience, particularly in the MC interface for precision tasks.
\textbf{H2:} Force feedback had a more significant impact on performance and user experience in the MC interface compared to force feedback in the KT interface.
\textbf{H3:} Force feedback showed more pronounced benefits for complex tasks requiring precision, with only minor improvements for simpler tasks.

% These results underscore the importance of selecting appropriate force feedback tailored to both task complexity and the type of interface.

\section{Conclusion}

% \textcolor{red}{rewrite}

% Our results indicate that while force feedback can significantly enhance user performance and experience, its impact on data quality is not consistent. The interaction methods, the type of force feedback and the complexity of the tasks play a key role in these effects. These results contribute to the design of more effective robot control interfaces for robot data collection and teleoperation.

This study explored the impact of force-feedback Human-Robot Interaction interfaces for robot data collection. A user study with 31 participants examined how force feedback influences performance and user experience. While force feedback generally enhanced both aspects, its effects varied based on the interaction method, feedback type, and task complexity.
Additionally, we introduced an innovative integration of force feedback with Kinesthetic Teaching using XR. Its consistently high performance highlights its potential for more intuitive and effective interactions. These findings emphasize the need for customized force feedback designs, encouraging researchers to refine them for specific application requirements.

% In this study, we investigate the impact of force feedback Human-Robot Interaction interfaces in Learning from Demonstration. A user study with 31 participants was conducted to examine how force feedback affects performance and user experience. Force feedback improved both user experience and performance, but its effects varied depending on the interaction method, type of force feedback, and task complexity. Additionally, we developed a novel combination of force feedback with Kinesthetic Teaching and XR.

% Force feedback improved both user experience and performance, but its effects varied depending on the interaction method and interface. In the MC interface, where users interact with a virtual robot at a distance, force feedback provides clear haptic cues, improving performance and user experience, especially in tasks requiring repositioning. In contrast, the KT interface, with direct physical interaction, showed slight benefits from force feedback, as visual cues were already sufficient.
% The impact of force feedback also depended on task complexity. For precision-demanding tasks, force feedback significantly improved success rates, efficiency, and user satisfaction. In simpler tasks, the improvements were minor and not statistically significant.

% Overall, force feedback can enhance interaction performance and user experience in  HRI, but its impact varies with the interaction method and task complexity. 

\bibliographystyle{IEEETran}
\bibliography{references}
\clearpage

\end{document}